\pdfoutput=1

\documentclass[11pt]{article}

\usepackage[preprint]{acl}

\usepackage{times}
\usepackage{latexsym}

\usepackage{amsmath}
\usepackage{algorithm}
\usepackage{algpseudocode}
\usepackage{amsfonts}
\usepackage{array} 
\usepackage{graphicx} 
\usepackage{makecell}
\usepackage[T1]{fontenc}

\usepackage[utf8]{inputenc}

\usepackage{microtype}

\usepackage{inconsolata}

\usepackage{graphicx}
\usepackage{booktabs}
\usepackage{amsmath}
\usepackage{enumitem}
\newcommand{\ie}{\emph{i.e., }}
\newcommand{\eg}{\emph{e.g., }}

\newcommand{\cf}{\emph{cf. }}

\title{Dual-Phase Accelerated Prompt Optimization}

\author{
Muchen Yang\textsuperscript{1}, 
~Moxin Li\textsuperscript{2}\thanks{\hspace{2mm}Corresponding Author},
~Yongle Li\textsuperscript{3}, 
~Zijun Chen\textsuperscript{3}, \\
\textbf{
~Chongming Gao\textsuperscript{1}\footnotemark[1],
~Junqi Zhang\textsuperscript{4},
~Yangyang Li\textsuperscript{5},
~Fuli Feng \textsuperscript{1,3}
}
\\
\textsuperscript{1}University of Science and Technology of China, 
~\textsuperscript{2}National University of Singapore\\
\textsuperscript{3}Institute of Dataspace, Hefei Comprehensive National Science Center \\
~\textsuperscript{4}AtomEcho Inc.,
~\textsuperscript{5}Academy of Cyber
\\
\tt{muchen00@mail.ustc.edu.cn},
\tt{limoxin@u.nus.edu},
~\tt{liyongle999@gmail.com}\\
~\tt{zijunchen248@gmail.com},
\tt{chongminggao@ustc.edu.cn},
~\tt{zhangjunqi@atomecho.xyz}\\ 
~\tt{liyangyang@live.com},
~\tt{fulifeng93@gmail.com}\\
}

\begin{document}
\maketitle

\begin{abstract}
Gradient-free prompt optimization methods have made significant strides in enhancing the performance of closed-source Large Language Models (LLMs) across a wide range of tasks. 
However, existing approaches make light of the importance of high-quality prompt initialization and the identification of effective optimization directions, thus resulting in substantial optimization steps to obtain satisfactory performance.
In this light, we aim to accelerate prompt optimization process to tackle the challenge of low convergence rate. 
We propose a dual-phase approach which starts with generating high-quality initial prompts by adopting a well-designed meta-instruction to delve into task-specific information, and iteratively optimize the prompts at the sentence level, leveraging previous tuning experience to expand prompt candidates and accept effective ones.
Extensive experiments on eight datasets demonstrate the effectiveness of our proposed method, achieving a consistent accuracy gain over baselines with less than five optimization steps. 

\end{abstract}

\section{Introduction}

LLMs have demonstrated remarkable capabilities across a wide range of natural language processing (NLP) tasks, including machine translation~\cite{Qin2024LargeLM}, summarization~\cite{Goyal2022NewsSA}, and question answering~\cite{Zhang2023ExtractiveSV}.
The dependency on prompt quality has led to the emergence of prompt engineering \cite{Diao2023ActivePW, White2023APP}, aiming at crafting effective prompts to elicit the desired responses from LLMs. 
As the need for efficient prompt design becomes increasingly evident \cite{Liu2021GPTUT}, automatic prompt optimization has been introduced to streamline the prompt design process, ensuring that LLMs are utilized to their full potential \cite{Gao2021MakingPL,Liu2021PretrainPA,Reynolds2021PromptPF}.

\begin{figure}
    \centering
    \includegraphics[width=1\linewidth]{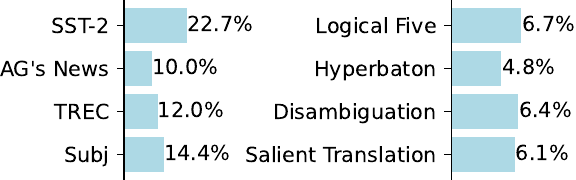}
    \vspace{-4mm}
    \caption{Average accuracy improvement on eight datasets with \emph{four} optimization steps.}
    \label{fig:intro}
    \vspace{-3mm}
\end{figure}

Automatic prompt optimization can be broadly categorized into gradient-based and gradient-free methods. 
Gradient-based methods \cite{Shin2020ElicitingKF,Li2021PrefixTuningOC,Liu2021GPTUT,Liu2022PTuningPT} are devised for open-source LLMs to enable the optimization of prompts through adjustments based on model gradient.
Gradient-free methods have emerged as the predominant approach for closed-source LLMs, which focuses on refining prompts without access to the model gradient \cite{Prasad2022GrIPSGE,Yang2023LargeLM,Guo2023ConnectingLL}. 
Starting from initial prompts, these methods usually expand candidate prompts using searching methods \cite{Pryzant2023AutomaticPO,Wang2023PromptAgentSP} and then accepting the more prominent ones in an iterative manner. This paper focuses on gradient-free methods due to the distinguished abilities of closed-source LLMs and the challenge of optimizing their prompts with limited model information.

We argue that current gradient-free prompt optimization methods have not adequately considered the rate of convergence. 
Typically, these methods demand an excessive number of optimization steps to obtain satisfactory prompts due to the limited access to model details, the vast discrete search space, and the uncertain optimization directions~\cite{Wang2023PromptAgentSP,Pan2023PlumPL,Yang2023LargeLM}. 
Representative work such as OPRO \cite{Yang2023LargeLM} even necessitates nearly 200 optimization steps for some NLP tasks.  
This requirement for excessive optimization steps makes existing methods impractical for real-world applications since users are understandably reluctant to tolerate extensive optimization steps to achieve satisfactory performance levels.
Therefore, we aim to achieve accelerated prompt optimization, obtaining satisfactory performance via few optimization steps (\eg $<5$).

To achieve accelerated prompt optimization, two crucial factors need to be considered: high-quality initial prompts and effective optimization directions. 
Firstly, the initialization of the prompt plays a crucial role in determining the efficiency
of the optimization process \cite{Ye2023PromptEA}, whereas existing approaches pay insufficient attention to the impact of initialization on subsequent optimization. 
Therefore, we aim to obtain initial prompts of high quality, laying a solid foundation to accelerate optimization process. 
Secondly, the accelerated prompt optimization needs to identify the most effective optimization directions in each step, streamlining efficient optimization from the initial prompts. 
Thus, we aim to design a more refined expansion tuned by experience and acceptance of candidate prompts enhanced by examination of failure cases.

To this end, we propose a dual-phase approach to achieve the accelerated gradient-free prompt optimization. 
Our approach consists of two phases: high-quality initial prompt generation, and 
experience-tuned optimization. 
Firstly, we utilize a well-designed meta-instruction to guide the LLM in generating high-quality and structured initial prompts that contain task-specific information, including task type and description, output format and constraints, suggested reasoning process, and professional tips. 
After that, we devise a sentence-level prompt optimization strategy for efficiently optimization on the long initial prompt, leveraging previous direction tuning experience, together with failure cases, to select sentences in the initial prompt to be expanded and accept effective prompt candidates.
Extensive experiments (\cf Figure~\ref{fig:intro}) on three LLMs across several datasets confirm the effectiveness and superiority of our method. Our contributions are threefold:
\begin{itemize}[leftmargin=*]
    \item We reveal the issue of low convergence rate in gradient-free prompt optimization, and highlight the problem of accelerated prompt optimization.
    \item We propose a dual-phase approach, achieving accelerated prompt optimization through high-quality initial prompt generation and experience-tuned optimization. 
    \item We conduct extensive experiments, demonstrating that the proposed method achieves satisfying performance within few optimization steps.
\end{itemize}
\section{Related Work}
The gradient-free prompt optimization for closed-source LLMs typically contains two phases: initialization and iterative optimization steps, where the optimization step consists of expansion and selection stages. 

\paragraph{Initialization.}
The prompt initialization for optimization can be achieved manually or autonomously.
Manual initialization often entails professional machine learning engineers formulating prompts, as delineated in \cite{Pryzant2023AutomaticPO}. Concurrently, works such as \cite{Guo2023ConnectingLL}, \cite{Pan2023PlumPL}, and \cite{Wang2023PromptAgentSP} utilize existing manual prompts as the foundational set to harness human creativity. 
In contrast, automated initialization leverages the power of LLM generation, which is exemplified by \cite{Zhang2023AutoInstructAI}, generating prompts from few-shot exemplars and a rudimentary description, and \cite{Zhou2022LargeLM}, fabricating prompts based on meta-prompts and illustrative input-output examples.
Our method belongs to the automated initialization, improving the initial prompt generation for acceleration. 

\paragraph{Optimization.}
The optimization step is achieved by expanding prompt candidates by modifying from the initial prompt and selecting the better candidates for the next iteration. 
The expansion stage can be executed through rephrasing, as in \cite{Zhou2022LargeLM}, where high-scoring prompts undergo evolution akin to a Monte Carlo search methodology, or through heuristic algorithms that automatically revise prompts, as in \cite{Guo2023ConnectingLL} and \cite{Pan2023PlumPL}. More complex regeneration strategies are employed by works like \cite{Wang2023PromptAgentSP}, where the optimizer LLM progressively expands prompts based on task delineations and historical iterations. 
The expansion can also be implemented leveraging an open-source LLM \cite{Lin2023UseYI, Chen2023InstructZeroEI}. 
Reinforcement learning-based methods have also been adopted for prompt modification \cite{diao2023blackbox}. 
Moreover, the granularity of prompt modification exhibits variation across studies. Heuristic-based methods and \cite{Hsieh2023AutomaticEO} work operate at the word/token granularity, while classical optimization algorithms like \cite{Pryzant2023AutomaticPO, Zhou2022LargeLM} consider the entire prompt. 
The selection stage generally utilized the performance of the prompt on a held-out validation set \cite{Pryzant2023AutomaticPO, Zhou2022LargeLM, Wang2023PromptAgentSP}, while recent work also explores human preference feedback \cite{Lin2024PromptOW} or score feedback from other LLMs \cite{yang2024large}. 

\begin{figure*}[t]
    \centering
    \includegraphics[width=0.95\linewidth]{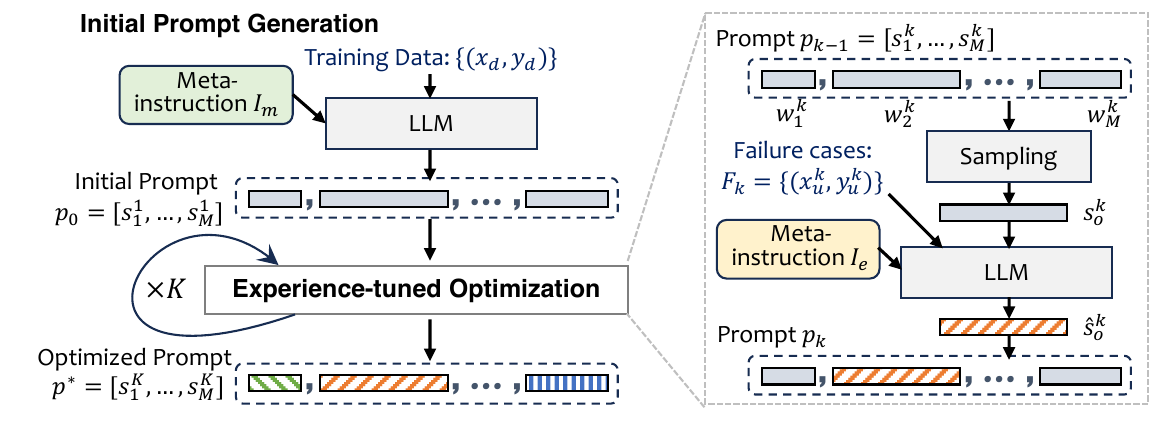}
    \caption{Illustration of the proposed method.}
    \label{fig:method}
\end{figure*}

\section{Problem Formulation}

\subsection{Gradient-Free Prompt Optimization}
For a target NLP task $\mathcal{T}$ with input $x$, the closed-source LLM predicts the output $\hat{y}$ given $x$ concatenated with the prompt $p$, where $x, \hat{y}$ and $p$ are all word sequences.
The aim for prompt optimization is to find an optimal prompt $p^*$ that obtains the desired $\hat{y}$, which can be evaluated by metrics such as accuracy with reference to the ground truth $y$. 
The gradient-free prompt optimization contains an initialization phase followed by $K$ iterative optimization steps. The $k$-th optimization step starts from an initial prompt $p_{k-1}, k \in [1, K]$, and sequentially performs two stages: expansion of prompt candidates, and acceptance of the prominent prompts as the next initial prompts, as detailed below.

\paragraph{Expansion of Prompt Candidates.}
At the $k$-th optimization step, The expansion stage search for new prompt candidates with potential better performance starting from $p_{k-1}$, with searching methods 
such as edit-based \cite{Prasad2022GrIPSGE} and LLM rewriting \cite{Pryzant2023AutomaticPO}. 
Formally, the expansion function $f_E(\cdot)$ generates prompt candidate set $P^c_k = \{p^c_{k_1}, \cdots, p^c_{k_Q} \}$ with size $Q$.
\begin{align}
    P^c_k = f_E(p_{k-1}). 
\end{align}

\paragraph{Acceptance of Prominent Prompts.}
The acceptance stage evaluates the performance of each prompt candidate in $P^c_k$ to determine whether it should be continued for next optimization step. This is usually achieved by evaluation on a held-out validation set $V = \{(x^v, y^v)\}$, and accepting the top-performing prompt candidates. Formally, with the evaluation function on LLM as $f_S(\cdot)$,
\begin{align}
    r^k_i &= f_S(p^c_{k_i}, V), i\in[1, \cdots, Q], \\ \notag
    p_{k} &= p^c_{k_j}, \text{where } j = \text{argmax}(\{r^k_1, ..., r^k_Q\}). 
\end{align}
where $\text{argmax}(\cdot)$ denotes the index of the maximum value. 
At the final optimization step, the top-performing prompt $p_{K}$ will be accepted as the optimized prompt $p^*$.

\subsection{Accelerated Prompt Optimization}
Although current research on gradient-free prompt optimization can achieve significant performance gains on multiple tasks, demands for a great number of optimization steps hinder their practicability in real-world scenarios. 
For instance, \citet{Yang2023LargeLM} does not converge even after over 150 steps in some tasks; \citet{Wang2023PromptAgentSP} finds a good solution in 
50 to 75 steps. 
Therefore, we highlight the problem of accelerated prompt optimization, \ie obtaining $p^*$ with satisfactory performance in few optimization steps, \eg $K<5$.

\section{Proposed Method}

\subsection{Motivation}
We believe that two factors are crucial for achieving accelerated prompt optimization, which current gradient-free prompt optimization methods fail to achieve.
Firstly, the initial prompt $p_0$ plays a crucial role in accelerating the prompt optimization process \cite{Ye2023PromptEA}, where $p_0$ with better LLM performance makes the optimization towards better prompts easier, preventing LLMs from excessively exploring suboptimal prompt regions. 
This is generally overlooked by existing research that utilizes uninformative initial prompts, \eg \cite{Yang2023LargeLM}. Therefore, we propose to devise high-quality $p_0$ by crafting a novel initial prompt schema. 
Furthermore, a more precise expansion and acceptance of prompt candidates ensure highly efficient optimization direction and fewer optimization steps. 
Current expansion and acceptance techniques optimize the prompt towards improving the general task performance, where effective optimization direction in each step is hard to ensure. 
To tackle this, we propose to utilize the past failure cases from previous optimization steps to further navigate the expansion and acceptance of prompt candidates. 
We illustrate our dual-phase approach as follows (\cf Figure~\ref{fig:method}).  

\subsection{High-Quality Initial Prompt Generation}
\begin{figure}[t]
    \centering
    \includegraphics[width=1\linewidth]{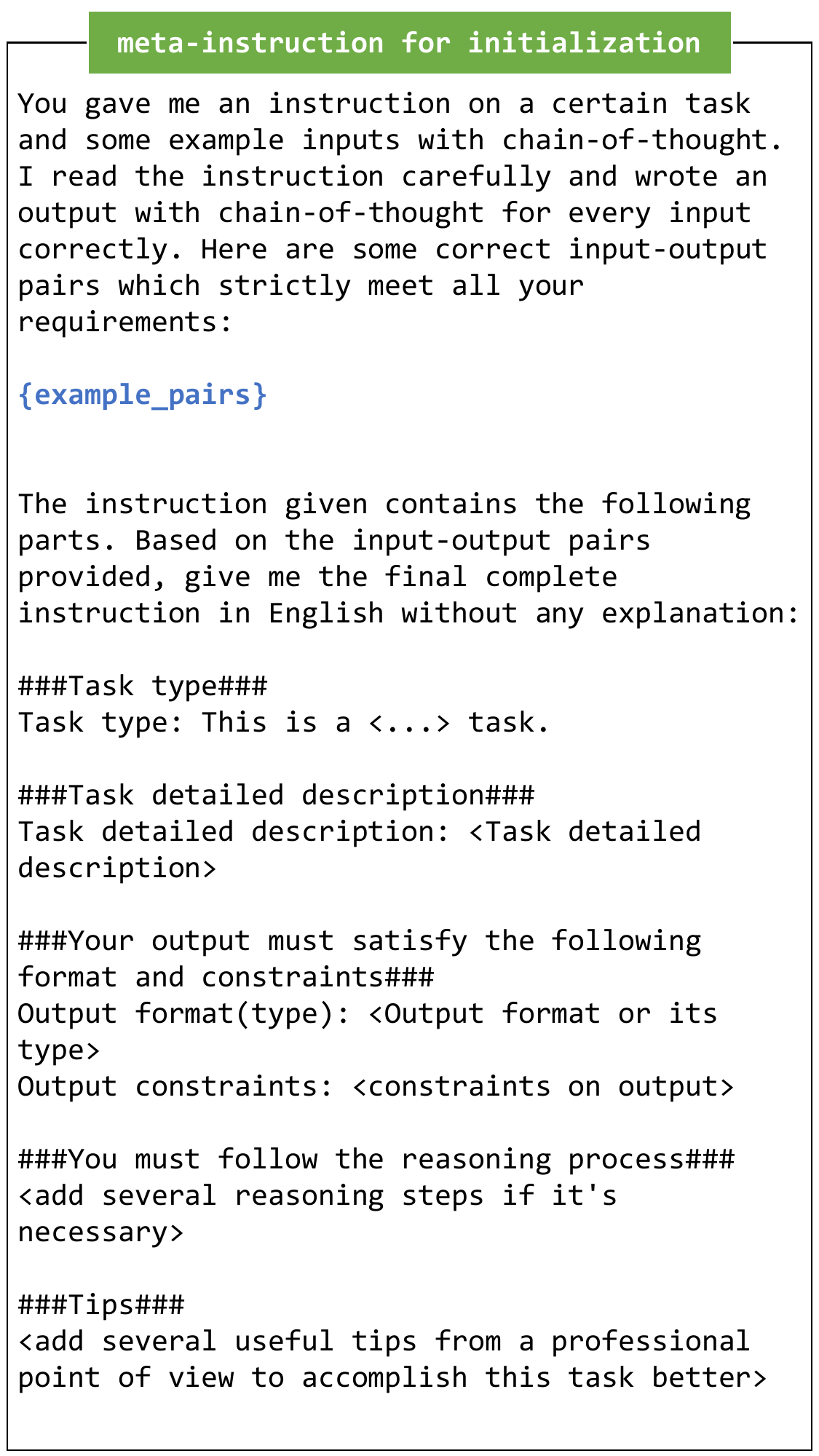}
    \caption{Meta-instruction used in our initialization phase to generate high-quality initial prompts.}
    \label{fig:meta-initial}
\end{figure}

We think that a high-quality initial prompt that can elicit the desired response from LLMs should be able to provide clear task instruction and detailed task-related information. Specifically, it should
1) give a clear definition of the task type and provide a detailed task description, 
2) define the output format and constraints, 
3) provide insights on the reasoning processes and professional tips.
To achieve such initial prompts, we are inspired by the step-back prompting \cite{Zheng2023TakeAS} which demonstrates LLM's ability to derive high-level concepts and principles from examples. 
Thus, following \cite{Zhou2022LargeLM}, we design a meta-instruction $I_m$ (\cf Figure~\ref{fig:meta-initial}), leveraging LLM's ability to generate $p_0$ by observing the input-output exemplars of the target task $\mathcal{T}$ and inferring the above required information. Formally, defining 
input-output exemplars as $D = \{(x_d, y_d)\}$, 
\begin{align}
    p_0 = LLM(I_m, D). \label{eq:init_p0}
\end{align}

\subsection{Experience-Tuned Optimization}
In the optimization phase, it is necessary to tune the expansion and acceptance of prompt candidates to quickly improve the task performance as evaluated on the validation set $V$ and thus reduce optimization steps. 
Inspired by previous research \cite{Pryzant2023AutomaticPO}, we intend to make the best of past failure cases to generate promising prompt candidates and filter out unnecessary optimization attempts.
In each optimization step, we maintain a failure case set $F_k = \{(x_k^f, y_k^f)\}$ containing the examples from $V$ where the initial prompt $p_{k-1}$ fails to predict the ground truth in the acceptance stage, \ie $\hat{y}_k^f \neq y_k^f$.

\paragraph{Expansion.}
In the expansion stage, since the initial prompts are long prompts with at least four sentences, we aim to improve the expansion efficiency by segmenting them into individual sentences for sentence-level expansion following LongPO \cite{Hsieh2023AutomaticEO}. 
Moreover, since different sentences in the initial prompts contain different task-related information and may have different impacts on the task performance, we devise sentence weights $w^k$ to estimate the impact of each sentence on the performance improvement, 
which is updated leveraging the past failure cases. 
We first split the initial prompt $p_0$ into $M$ sentences, and initialize the weight $w^1$ for each sentence as 1. 
\begin{align}
    p_0  &= [s^1_1,s^{1}_2,...,s^{1}_{M} ], \\ \notag
    w^1_t & = 1, t \in [1, M]. 
\end{align}
In the $k$-th optimization step, we compute the acceptance probability $\text{Pr}^k$ for each sentence: 
\begin{align}
\label{eq:selection_prob}
     \text{Pr}^k_i &= \frac{\exp(w^{k}_i)}{\sum_{j=1}^{M} \exp(w^k_j)}. 
\end{align}

\begin{figure}[t]
    \centering
    \includegraphics[width=1\linewidth]{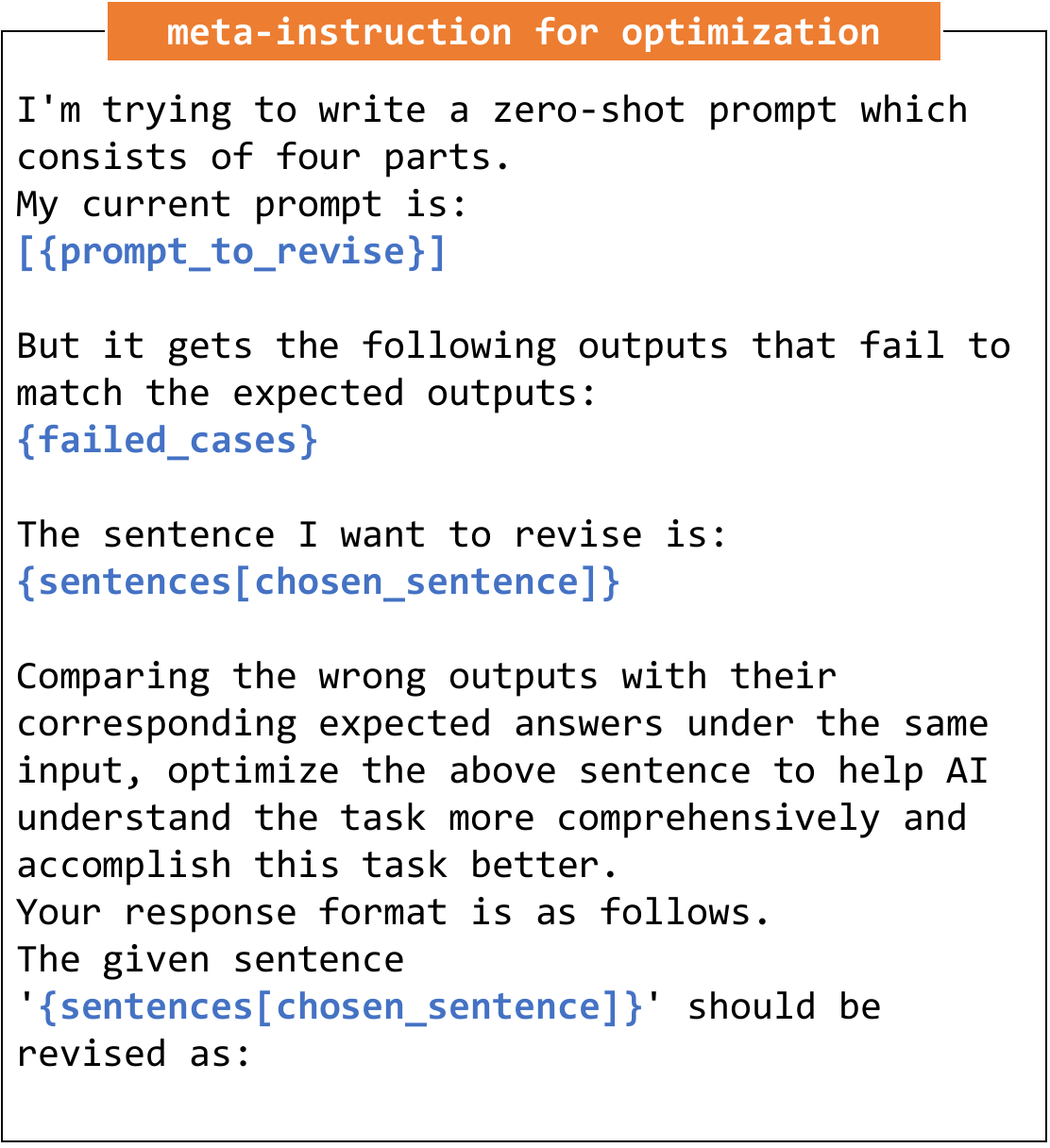}
    \vspace{-2mm}
    \caption{Meta-instruction used in the optimization phase.}
    \label{fig:meta-optimize}
    \vspace{-2mm}
\end{figure}

After that, we sample a sentence for expansion based on the probability distribution $\text{Pr}^k = [\text{Pr}^k_1, \cdots, \text{Pr}^k_{M}]$, where the sampled sentence is denoted as $s^k_o, o \in [1, M]$. 
For expansion of $s^k_o$, we 
design a meta-instruction $I_e$ (\cf Figure~\ref{fig:meta-optimize}) to instruct LLM to generate a revised sentence considering the past experience. 
\begin{align}
    \label{eq:meta_optimize_generation}
    \hat{s}^k_o = LLM(I_e, p_{k-1}, F_k, s^k_o).
\end{align}

Before passing $\hat{s}^k_o$ to the acceptance stage, we design additional strategies to further guarantee the effectiveness of the generated sentence leveraging $F_k$. 
Firstly, to ensure $\hat{s}^k_o$ can actually improve over $s^k_o$, we replace $s^k_o$ in $p_{k-1}$ with $\hat{s}^k_o$, denoted as $\hat{p}_k$, and evaluate whether $\hat{p}_k$ outperforms $p_{k-1}$ on $F_k$. 
We accept $\hat{s}^k_o$ only when $\hat{p}_k$ has improved the performance over $p_{k-1}$ larger than a threshold $H_F$. 
\begin{align}
    f_S(\hat{p}_k, F_k) - f_S(p_{k-1}, F_k) > H_F. \label{eq:H_F}
\end{align}
Besides, to avoid repeatedly generating the same ineffective $\hat{s}^k_o$, we build a collection $\mathcal{G}$ of undesired sentence revisions and check whether $\hat{s}^k_o$ has appeared in $\mathcal{G}$. If the above two criteria are not met, we abandon $\hat{s}^k_o$ and regenerate starting from Eq.~\ref{eq:meta_optimize_generation}.

\paragraph{Acceptance.}

In addition to evaluating $\hat{p}_k$'s performance on the entire failure case $F_k$, we also evaluate its performance on the validation set $V$.
We accept $\hat{p}_k$ as the next initial prompt $p_k$ only when $\hat{p}_k$ has improved the performance over $p_{k-1}$ larger than a threshold $H_V$. Otherwise, we abandon $\hat{p}^k$ and restart from sampling $s^k_o$. 
\begin{align}
\label{eq:H_V}
    f_S(\hat{p}_k, V) - f_S(p_{k-1}, V) > H_V.  
\end{align}
If $\hat{p}^k$ is accepted, we update its sentence weights.
We calculate the mixed evaluation result $f_R(\cdot)$ and update the $w^{k+1}$ as follows, where $\alpha$ and the learning rate $\eta$ are adjusting hyperparameters. 
\begin{align}
    f_R(\hat{p}_k) & = \alpha f_S(\hat{p}_k, V) + (1-\alpha) f_S(\hat{p}_k, F_k). \\ \notag
    w^{k+1}_i & = w^k_i \exp(\frac{\eta f_R(\hat{p}_k)}{\text{Pr}^k_i M}).
\end{align}
When the number of times that Eq.~\ref{eq:H_F} or Eq.~\ref{eq:H_V} is not satisfied
accumulates to 5, we consider the algorithm to have converged.

The weight formula is designed to adaptively update the importance of each sentence in the prompt based on its impact on overall performance improvement. $f_R(\hat{p}_k)$ modulates the magnitude of the weight adjustment: a higher $f_R(\hat{p}_k)$ leads to larger updates. $\text{Pr}^k_i$ determines the weight's contribution, while $M$ is used for normalization to ensure balanced weight updates. The learning rate $\eta$ controls the extent of weight adjustments based on the evaluation feedback. Inspired by the EXP3 algorithm \cite{Auer1995GamblingIA}, these components facilitate a dynamic and adaptive optimization process, tuned by empirical performance data.
The who process is summarized in Algorithm~\ref{algo}.

\begin{algorithm}
\centering
\caption{\\ Dual-Phase Accelerated Prompt Optimization}
\label{algo}
\begin{algorithmic}[1]
\Require Input-output exemplars $D$, validation set $V$, 
meta-instruction $I_m$ and $I_e$.
\Ensure Optimized prompt $p^*$
\State Initialize $p_0$ (Eq.~\ref{eq:init_p0}),  derive failure case set $F_1$
\State Split $p_0$ into $M$ sentences $[s^1_1, s^1_2, \ldots, s^1_M]$, initialize sentence weights $\{w^1_i\}_{i=1}^M \gets 1$, $k \gets 1$
\While{not converged}
    \State $\triangleright{\textit{\textbf{  Expansion}}}$
    \State Sample a sentence $s^k_o$ based on $\text{Pr}^k$ (Eq.~\ref{eq:selection_prob}) \label{Sample a sentence}
    \State Generate revised sentence $\hat{s}^k_o$ (Eq.~\ref{eq:meta_optimize_generation})\label{Generate revised sentence}
    \State Replace $s^k_o$ in $p_{k-1}$ with $\hat{s}^k_o$ to get $\hat{p}_k$
    \If{$\hat{s}^k_o \in \mathcal{G}$ \textbf{or} (Eq.~\ref{eq:H_F}) is not satisfied}
        \State Add $\hat{s}^k_o$ to $\mathcal{G}$
        \State Regenerate $\hat{s}^k_o$ from line \ref{Generate revised sentence}
    \EndIf
    
    \State $\triangleright{\textit{\textbf{  Acceptance}}}$
    \If{(Eq.~\ref{eq:H_V}) is not satisfied}
        \State Restart from line \ref{Sample a sentence}
    \EndIf
    \State $p_{k} \leftarrow \hat{p}_k$, update $w^{k+1}_i$, $k \gets k+1$
    \State Update $F_{k}$ with new failure cases
\EndWhile
\State \Return optimized prompt $p^*=p_{k}$
\end{algorithmic}
\end{algorithm}
\vspace{-2mm}
\section{Experiments}
In this section, we begin by detailing datasets, baselines, and the implementation of the experiments. Following this, we conduct comprehensive and controlled experiments on our method.

\subsection{Experimental Settings}

\paragraph{Datasets.}
Our experiments are first conducted on general natural language understanding tasks across four datasets to validate our method, specifically focusing on sentiment classification (SST-2 \cite{Socher2013RecursiveDM}), topic classification (AG’s
News \cite{Zhang2015CharacterlevelCN}, TREC \cite{Voorhees2000BuildingAQ}) and subjectivity classification (Subj \cite{Pang2004ASE}). 
Then we perform our approach to the challenging BBH tasks \cite{Suzgun2022ChallengingBT}, 
which include manually provided few-shot Chain-of-Thought (CoT) prompts containing task descriptions and demonstrations.

\paragraph{Baselines.}
We compare our method with three popular prompt optimization methods for zero-shot black-box prompting and the well-crafted prompts manually provided in BBH tasks:
    \textbf{APO} \cite{Pryzant2023AutomaticPO}: 
    Generating natural language ``gradients'' to criticize and improve the current prompts.
    \textbf{APE} \cite{Zhou2022LargeLM}: Proposing both a naive and an iterative Monte Carlo search methods to approximate the solution to the prompt optimization problem.
    \textbf{PromptAgent} \cite{Wang2023PromptAgentSP}: 
    Automating expert-level prompt generation by treating it as a strategic planning problem using Monte Carlo tree search and error feedback to refine and optimize prompts.
    \textbf{Manual Prompt} \cite{Suzgun2022ChallengingBT}: The few-shot CoT version of human-designed prompts with teaching examples developed in BBH tasks.

\paragraph{Implementation Details.}
In line with \cite{Wang2023PromptAgentSP}, since BBH tasks lack an official train-test split, we shuffle the data and allocate approximately half for testing. The rest is used for training, prompt generation, and optimization. For datasets with predefined test sets, we use those directly.

Unless otherwise stated, we evaluate performance (\ie accuracy) on GPT-3.5-Turbo using the OpenAI API\footnote[1]{\url{https://chat.openai.com/}} (currently gpt-3.5-turbo-0125) in a zero-shot prompt setting. The temperature is set to 0 for prediction and 0.5 for prompt generation to enhance diversity.
To accelerate prompt optimization, we limit the maximum optimization steps to \textbf{four} for all methods, while keeping other baseline parameters and settings at default. 
At the beginning of prompt initialization, eight exemplars are obtained by concatenating unique input-output pairs from the shuffled training data until the desired amount is reached, ensuring no duplicate inputs.
Due to limited computational resources, our approach generates and optimizes only one initial prompt.
By default, we set $H_F=0.3$, $H_V=0.1$, $\alpha=0.4$, and $\eta=0.055$ in Algorithm ~\ref{algo} to accelerate the optimization phase.

\subsection{Main Results \& Analysis}

\begin{table}[!th]
  \centering
  \footnotesize
  \tabcolsep=2pt
  \begin{tabular}{l|c|cccc}
    \toprule
    & \textbf{Few-shot} & \multicolumn{4}{c}{\textbf{Zero-shot}} \\
    \midrule
    \textbf{Task} & \textbf{Manual} & \textbf{APO} & \textbf{APE} & \textbf{PA} & \textbf{Ours} \\
    \midrule
    \textbf{SST-2} & \slash & 0.89 & \underline{0.92} & 0.443 & \textbf{0.978} \\
    \textbf{AG’s News} & \slash & \underline{0.88} & 0.819 & 0.785 & \textbf{0.928} \\
    \textbf{TREC} & \slash & \textbf{0.795} & 0.513 & 0.687 & \underline{0.785} \\
    \textbf{Subj} & \slash & \underline{0.64} & 0.593 & 0.494 & \textbf{0.72} \\
    \midrule
    \textbf{Logical Five} & 0.388 & 0.392 & 0.404 & \underline{0.443} & \textbf{0.48} \\
    \textbf{Hyperbaton} & 0.744 & 0.808 & \underline{0.865} & 0.823 & \textbf{0.88} \\
    \textbf{Disambiguation} & 0.580 & 0.688 & 0.645 & \underline{0.696} & \textbf{0.74} \\
    \textbf{Salient Translation} & 0.544 & 0.456 & \underline{0.538} & 0.468 & \textbf{0.548} \\
    \midrule
    \textbf{Avg.} & 0.564 & 0.694 & 0.662 & 0.605 & \textbf{0.757} \\
    \bottomrule
  \end{tabular}
  \caption{Accuracy on eight tasks on GPT-3.5-Turbo. PA indicates PromptAgent. Bold and underlined text indicate the best and second-best results, respectively.}
  \label{tab:main}
\end{table}

\begin{figure*}[!t]
    \centering
    \setlength{\abovecaptionskip}{0.1cm}
\setlength{\belowcaptionskip}{0cm}
    \includegraphics[width=0.9\textwidth]{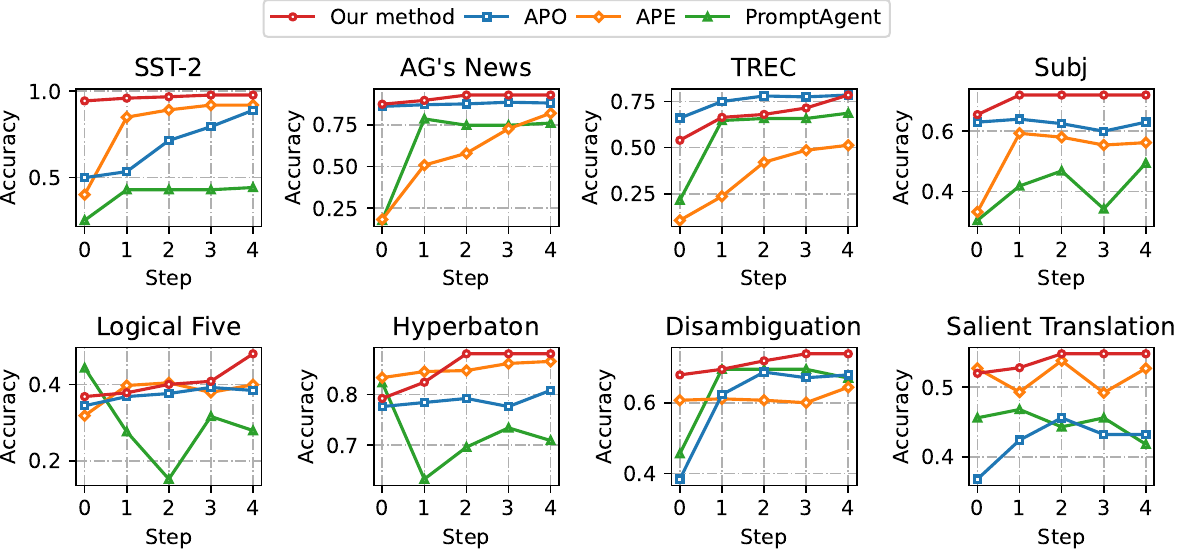}
    \caption{Performance (accuracy) over 4 steps across 8 tasks on GPT-3.5-Turbo.}
    \label{fig:gpt-all}
\end{figure*}

\paragraph{Overall Results.}
Table~\ref{tab:main} demonstrates the effectiveness of our accelerated dual-phase approach across 8 NLP tasks compared to classic prompt optimization methods. Our method significantly outperforms all baselines, achieving an average improvement of approximately \textbf{10.7\%} over APO, \textbf{16.4\%} over APE, and \textbf{29.7\%} over PromptAgent across the given tasks.

Our method also surpasses few-shot CoT human-crafted prompts with an approximately \textbf{17.6\%} average improvement on selected BBH tasks, indicating its ability to produce high-quality prompts that enhance the black-box LLM’s capabilities in logical deduction, grammar, language understanding, and multilingual tasks without teaching examples.

\paragraph{Analysis.}
To understand this result, we analyzed the prompt expansion and acceptance processes:
In prompt expansion, our method leverages past experience, filters out unnecessary optimization attempts, and collects undesired revisions. This contrasts with baseline methods that inefficiently explore prompt space and underutilize past iterations. APE lacks reflection on past iterations, slowing its Monte Carlo-based search. APO uses error feedback to guide beam search but is slowed by evaluating many paths. PromptAgent’s Monte Carlo Search Tree explores prompt optimization through simulations, but limited steps lead to suboptimal results.

In the acceptance process, inspired by the EXP3 algorithm, our method uses weighted sentences and modifications to enhance prompt quality, making it superior in identifying promising candidates and optimizing directions.

\paragraph{Convergence Analysis.}
To evaluate our method’s convergence within four steps compared to others, we examine how quickly each method achieves peak performance across datasets. Figure \ref{fig:gpt-all} shows the performance (accuracy) variation of four prompt optimization methods across eight datasets, with each subfigure representing a different dataset.
While APO, APE, and PromptAgent experience fluctuations or plateau at lower accuracy, our method demonstrates the fastest convergence across most datasets, often reaching near-peak performance within the first two steps. This rapid convergence highlights our method’s efficiency in optimizing prompts quickly and effectively, making it promising for tasks requiring prompt optimization within a few steps.

\subsection{Ablation Study}

We conduct several ablation experiments to assess the efficacy of our method.

\begin{figure}[!t]
    \centering
    \includegraphics[width=0.46\textwidth]{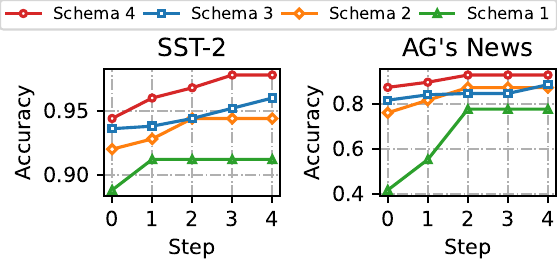}
    \caption{Results on GPT-3.5-Turbo with different initial prompt schemas.}
    \label{fig:ablation-init}
\end{figure} 

\begin{figure}[!t]
    \centering
    \includegraphics[width=0.46\textwidth]{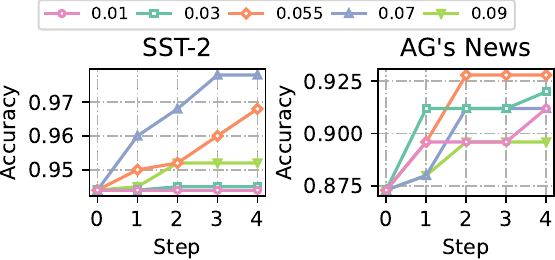}
    \caption{Results on GPT-3.5-Turbo with different optimization learning rates.}
    \label{fig:ablation-lr}
\end{figure}

\begin{figure*}[!t]
    \centering
    \includegraphics[width=0.95\textwidth]{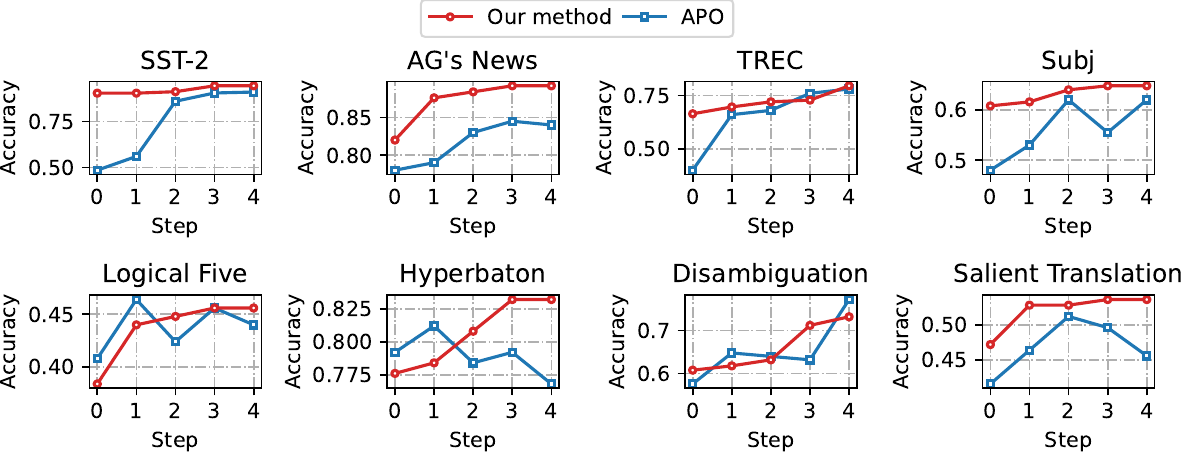}
    \caption{Accuracy over 4 steps across 8 tasks on Baichuan2-Turbo.}
    \label{fig:ablation-baichuan}
\end{figure*}

\begin{figure*}[!t]
    \centering
    \includegraphics[width=0.95\textwidth]{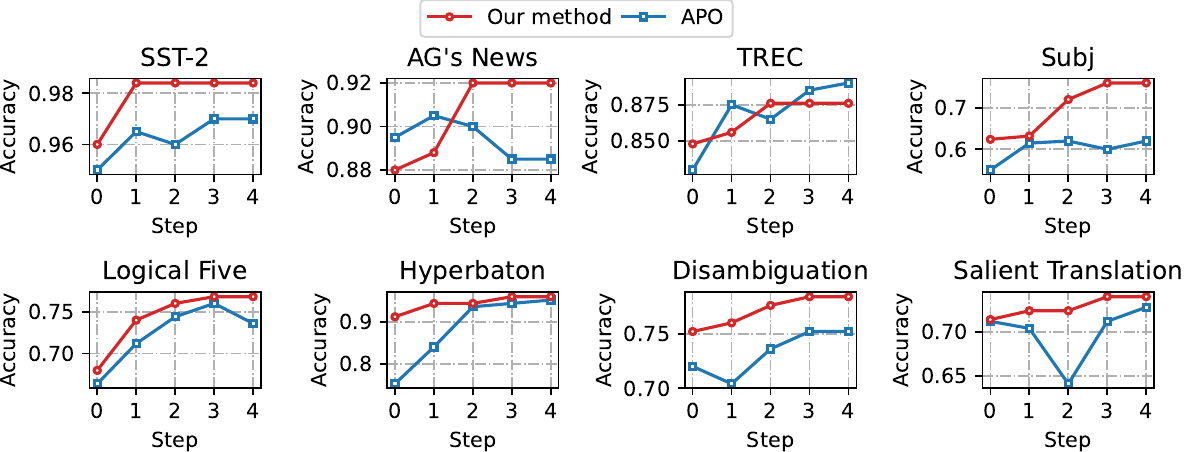}
    \caption{Accuracy over 4 steps across 8 tasks on GPT-4.}
    \label{fig:ablation-gpt4}
\end{figure*}

\subsubsection{Different Initial Prompt Schemas}

Our method uses a meta-instruction to generate a prompt with four components: 
a) task type and description,
b) output format and constraints,
c) suggested reasoning process, and
d) professional tips.
We define:
\emph{Schema 4}: All four components
\emph{Schema 3}: First three components
\emph{Schema 2}: First two components
\emph{Schema 1}: Task type and description only (common in current techniques).
We vary the meta-instructions for these schemas and conduct four-step prompt optimization experiments on SST-2 and AG’s News to assess their impact on optimization.

As shown in Figure \ref{fig:ablation-init}, initial prompts from Schema 4 yield the highest evaluation results. In contrast, Schema 1 has the lowest metrics and often falls into suboptimal local minima, a common issue with current methods. This comparison validates our meta-instruction design and underscores that a high-quality initial prompt is crucial for quickly identifying the optimal prompt.

\subsubsection{Sensitivity to Learning Rate}

During the optimization phase, the learning rate $\eta$ controls the extent of sentence weight updates after each round. A higher $\eta$ results in significant updates and responsiveness to recent performance changes, while a lower $\eta$ promotes stability with gradual adjustments. This balance is crucial for navigating the trade-off between exploration and exploitation.

We conduct prompt optimization experiments on SST-2 and AG’s News within four steps, testing $\eta$ values from 0.01 to 0.1. As shown in Figure \ref{fig:ablation-lr}, $\eta=0.055$ and $\eta=0.07$ are the most and second most effective in accelerating optimization.

\subsubsection{Performance on Different LLMs}

As Table \ref{tab:main} indicates, APO is the best baseline method. Therefore, we compare our method with APO using Baichuan2 \cite{Yang2023Baichuan2O} and GPT-4 accessed via the APIs. We conduct prompt optimization experiments on eight NLP datasets across four optimization steps. 

Figure \ref{fig:ablation-baichuan} and \ref{fig:ablation-gpt4} illustrate the performance variation of both methods across different datasets as optimization steps progress. APO fails to converge within four steps and shows greater performance volatility compared to Baichuan2-Turbo and GPT-4. In contrast, our method demonstrates rapid convergence and strong optimization acceleration. Except for the generalizability to other models, we also find that stronger LLM can achieve more effective prompt optimization with our method.

\subsubsection{Performance on Specialized Domain-Specific Task}

To evaluate our method performance on specialized tasks that require domain knowledge, we conduct experiments on the Geometric Shapes task \cite{Suzgun2022ChallengingBT}, which involves interpreting SVG paths to determine the geometric figures they represent, a task that requires specific domain knowledge.

\begin{table}[!t]
  \centering
  \small
  \tabcolsep=4pt
  \begin{tabular}{l|c|c}
    \toprule
    \midrule
    \textbf{Model} & \textbf{APO} & \textbf{Ours} \\
    \midrule
    \textbf{GPT-3.5-Turbo} & 0.36 & 0.392 \\
    \midrule
    \textbf{GPT-4} & 0.448 & 0.488 \\
    \bottomrule
  \end{tabular}
  \caption{Accuracy on Geometric Shapes task on GPT-3.5-Turbo and GPT-4.}
  \label{tab:Geo}
\end{table}

As shown in Table~\ref{tab:Geo}, our approach demonstrates consistent performance improvement over the best baseline APO, revealing the effectiveness of our method in specialized task.

\subsubsection{Results without Step Constraint}

\begin{figure}[!t]
    \centering
    \includegraphics[width=0.46\textwidth]{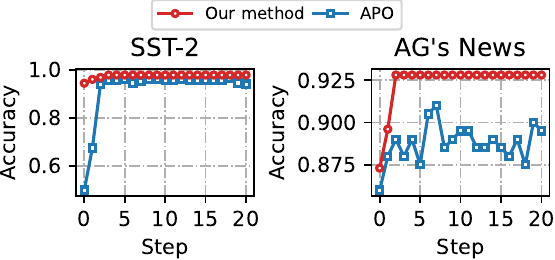}
    \caption{Accuracy over 20 steps on GPT-3.5-Turbo.}
    \label{fig:ablation-20step}
\end{figure}

We report the results of prompt optimization with a maximum of 20 steps on two general NLU tasks.
As shown in Figure \ref{fig:ablation-20step}, the strongest baseline, APO, converges on the SST-2 task with slightly lower accuracy than our method. However, on the AG’s News task, APO’s performance fluctuates significantly and lags behind our method.
Thus, our method demonstrates superior performance and faster convergence compared to existing methods, even with fewer optimization steps.

\subsubsection{Computational Complexity}
Since the running time is related to the number of API calls and may be affected by the network condition, we mainly present the number of API calls, which is an important metric for cost comparison on black-box LLMs.

\begin{table}[!th]
  \centering
  \small
  \tabcolsep=5pt
  \renewcommand\arraystretch{1.2}
  \begin{tabular}{l|c|c}
    \toprule
    \midrule
    \textbf{Task} & \textbf{APO} & \textbf{Ours} \\
    \midrule
    \textbf{SST-2} & 12,520 & 1,708  \\
    \textbf{AG’s News} & 12,733 & 2,089 \\
    \textbf{TREC} & 9,739 & 1,486 \\
    \textbf{Subj} & 12,790 & 1,848 \\
    \midrule
    \textbf{Logical Five} & 9,631 & 1,512 \\
    \textbf{Hyperbaton} & 9,934 & 1,626 \\
    \textbf{Disambiguation} & 9,471 & 1,187 \\
    \textbf{Salient Translation} & 10,190 & 1,451 \\
    \textbf{Geometric Shapes} & 9,648 & 1,496 \\
    \midrule
    \textbf{Avg.} & 10,739 & 1,600 \\
    \bottomrule
  \end{tabular}
  \caption{API calls consumed on nine tasks on GPT-4.}
  \label{tab:API calls}
\end{table}

 We conduct our experiments with GPT-4 on nine tasks. As shown in Table~\ref{tab:API calls}, our method requires approximately 1/7 of the number of API calls compared to the strongest baseline method, APO.
\section{Conclusion}
In this paper, we addressed the issue of low convergence rates in gradient-free prompt optimization methods for LLMs. 
Our proposed dual-phase approach effectively accelerates prompt optimization by generating high-quality initial prompts and leveraging tuning experience to navigate the optimization process. 
Extensive experiments on several LLMs across diverse datasets demonstrated the superiority of our method in achieving satisfactory performance within few optimization steps. 
Our approach not only enhances the efficiency of prompt optimization but also improves the overall performance of LLMs in various NLP tasks. 
Future work will focus on further refining the optimization strategies and exploring their applications in more diverse and complex scenarios.
\section*{Acknowledgements}

This work is supported by the National Natural Science Foundation of China (62272437,62402470), the University Synergy Innovation Program of Anhui Province (GXXT-2023-071), and the Fundamental Research Funds for the Central Universities of China (WK2100000053, PA2024GDSK0107). This research is also supported by the advanced computing resources provided by the Supercomputing Center of the USTC.

\clearpage
\newpage
\section*{Limitations}
We acknowledge some limitations despite the promising results of our research that could pave the way for future studies:

1) Our experiments were limited to general NLP tasks and one domain-specific task, more performance assessment on specialized tasks remains to be included.
2) Our method relies on labeled task data for prompt generation and evaluation, raising concerns about its robustness in personalized or scenarios lacking labeled data.
3) Our experiments were confined to GPT-3.5-Turbo, Baichuan2-Turbo and GPT-4, leaving the effectiveness of our method on other large language models to be validated in future studies.

Further study may be needed to address these limitations so as to improve the generalizability and robustness of our approach in broader and more complex real-world applications.

\bibliography{custom}

\end{document}